\documentclass[graybox]{svmult}

\usepackage{mathptmx}       
\usepackage{helvet}         
\usepackage{courier}        
\usepackage{type1cm}        
\usepackage{makeidx}         
\usepackage{graphicx}        
\usepackage{caption}   
\usepackage{subcaption}
\usepackage{multicol}        
\usepackage[bottom]{footmisc}
\usepackage{url}
\usepackage{multirow}

\makeindex             

\usepackage{xcolor}

\begin{document}

\title*{Development and Validation of Engagement and Rapport Scales for Evaluating User Experience in Multimodal Dialogue Systems}

\author{Fuma Kurata, Mao Saeki, Masaki Eguchi, Shungo Suzuki, Hiroaki Takatsu, and Yoichi Matsuyama}

%
%
\institute{Fuma Kurata \at Waseda University, Tokyo, Japan, \email{kurata@pcl.cs.waseda.ac.jp}}

\authorrunning{Fuma Kurata et al.}
\titlerunning{Development and Validation of Engagement and Rapport Scales}

\maketitle

\abstract{
This study aimed to develop and validate two scales of engagement and rapport to evaluate the user experience quality with multimodal dialogue systems in the context of foreign language learning. The scales were designed based on theories of engagement in educational psychology, social psychology, and second language acquisition.
Seventy-four Japanese learners of English completed roleplay and discussion tasks with trained human tutors and a dialog agent. After each dialogic task was completed, they responded to the scales of engagement and rapport. The validity and reliability of the scales were investigated through two analyses. We first conducted analysis of Cronbach's alpha coefficient and a series of confirmatory factor analyses to test the structural validity of the scales and the reliability of our designed items. We then compared the scores of engagement and rapport between the dialogue with human tutors and the one with a dialogue agent. The results revealed that our scales succeeded in capturing the difference in the dialogue experience quality between the human interlocutors and the dialogue agent from multiple perspectives.}

\section{Introduction}

In this study, we aim to develop and validate two scales of engagement and
rapport as the metrics of user experience quality with multimodal dialogue systems, in the context of foreign language learning.

Establishing valid evaluation metrics has been an important agenda in the Spoken Dialogue System (SDS) research as the guideline for developing and revising system components. Motivated by this, previous research has invented various approaches for evaluating the user's Quality of Experience (QoE), perhaps the most renowned is PARADISE framework\cite{walker1997paradise}. 
One of the advantages of the PARADISE framework is its high generalizability across various domains. 
Assuming that the notion of user satisfaction is relatively context-independent, this framework computes a holistic metric of user satisfaction as linearly predicted from two objectives of speaking: task success and dialogue costs. User satisfaction is estimated based on users' responses to post-dialogue questionnaires through a linear regression model. Once this linear model is trained, these objective measures can be used to automatically estimate user satisfaction.

Recently, dialogic tasks that SDS research needs to handle have been rapidly expanded.
Thanks to the development of Large Language Models (LLMs)\cite{deriu2021survey}, a variety of dialogic tasks can be easily designed even beyond the focus on task success and dialogue costs. In educational contexts, for instance, it is widely acknowledged that positive emotions during the interaction and social relationship with interlocutors (e.g., peers and teachers) play an integral role in facilitating students' learning processes\cite{xie2021conceptual}. 
In other words, to further extend SDS research, it is of paramount importance to evaluate user experiences not only from the perspective of task success and dialogue costs, but also from the perspective of their psychological state and social relationship with the system.

As metrics to measure these aspects, engagement has been used to assess the user's willingness to use\cite{ghazarian2020predictive}, and rapport has been employed to gauge trust and intimacy with the user\cite{zhao2014towards}. 
Challenges in measuring engagement and rapport lie in their conceptual and multidimensional nature. Moreover, to the best of our knowledge, there are no standardized methods for measuring these elements in the evaluation of dialogue systems. 
In human-agent interaction research, the definition of engagement varies across studies, and in some instances, a clear definition is not provided. Moreover, there is no consensus on the methods for measuring engagement \cite{glas2015definitions}. Similarly, scholars have suffered from the lack of valid measurement methods for rapport \cite{zhao2014towards}. Therefore we attempt to develop and validate methodological tools to capture user's subjective evaluations of dialogue systems in terms of engagement and rapport.
Specifically, we focus on evaluating a dialogue agent designed for English speaking learning\cite{saeki2021intella}. This agent acts as a tutor and facilitates learners' speaking learning through various tasks such as interviews, role-play, and discussions. Given the compatibility of engagement and rapport with reciprocal dialogues, we chose role-play and discussion tasks.

This study first formulated a conceptual model of engagement and rapport, drawing upon research in educational psychology and second language acquisition. 
Based on this model, we then designed a set of 21 questionnaire items. 
To evaluate the validity of the questionnaire items, we conducted a series of confirmatory factor analyses (CFA) to test the extent to which the items capture the proposed theoretical constructs. 
Additionally, this study aims to examine whether the questionnaires can differentiate the dialogue quality between experienced human tutors and dialogue systems. 
The research questions are formualted in this study as follows:
\begin{enumerate}
    \item To what extent do our designed items of engagement and rapport capture hypothesized structures of each construct?
    \item How do users' QoE measured with our designed items of engagement and rapport differ across interlocutor type (i.e., human vs virtual conversational agent)? 
\end{enumerate}
The paper is organized as follows: Section 2 reviews the theory of engagement and rapport. Section 3 describes the experimental methodology, specifically detailing the dialogue tasks, the participants in the experiment, the architecture of the dialogue system, the experimental design, the engagement and rapport questionnaires, and the data analysis methods. Section 5 discusses the results, and Section 6 concludes the paper.
\section{Engagement and Rapport}
\label{related:engagement_rapport}
In the fields of educational psychology and foreign language learning, engagement is a key concept in evaluating learner's participations. \cite{hiver2021engagement} defines engagement as "the amount (quantity) and type (quality) of learners’ active participation and involvement in a language learning task or activity". From the perspective of the importance of engagement, \cite{fredricks2004school} describes engagement as "what students do to further their learning" thereby highligting that the active involvement in their learning plays a vital role in their subsequent learning. Engagement in the learning process has been found to lead to many positive outcomes in education, such as continued learning, a sense of achievement, high learning motivation, and lower dropout rates \cite{hiver2021engagement, reschly2012jingle, christenson2012handbook}. Moreover, \cite{hiver2021engagement} identifies three characteristics of engagement: (1) it pertains to one's behavior; (2) it depends on the context and is influenced by the learning environment, human relationships, and learning tasks, and; (3) it always has an object, such as a person, topic, or task. It can thus be argued that engagement is formed not only by learner's attitude towards their learning but also by the interaction with external environments, including the people around them, the environment, and the nature of the tasks they undertake. 

To measure engagement comprehensively, one may need to consider the multidimensionality of engagement. According to \cite{hiver2021engagement}, engagement has four interrelated components: behavioural, cognitive, emotional, and social engagement. Behavioral engagement is concerned with learner's effort, quality of participation, and active involvement in learning tasks and challenges, all of which are observable as actions. Cognitive engagement reflects the mental effort of learners in the learning process, associated with intentional, selective, and sustained attention to achieve tasks and learning goals. Emotional engagement represents a variety of positive emotions such as enjoyment, enthusiasm, and anticipation, while negative emotions such as anxiety, boredom, and frustration are regarded as disengagement. This type of engagement can be a driving force for the other dimensions of engagement. Finally, social engagement is defined as the quality of social interactions between learners and their surrounding community. This dimension of engagement can be unique in that it is closely tied with peers and learning environments. 

Another important perspective of dialgoue experience quality may include rapport, which refers to a harmonious relationship between speakers and is associated with fun, connection, respect, mutual trust, and identification\cite{zhao2014towards}. 
The importance of rapport building in various contexts has been highlighted in the field of dialogue system research. Dialogue agents aimed at establishing rapport with users have been developed \cite{zhao2014towards, matsuyama2016socially}. These studies conceptualize rapport as consisting of three sub-dimensions: Face Management, Mutual Attentiveness, and Coordination. Face Management is defined as controlling one's behavior to be acknowledged or disapproved by others \cite{spencer2005politeness}, where politeness serves as a function to show acknowledgment towards the dialogue partner. Mutual Attentiveness is concerned with mutual attention and involvement with one another. They experience a sense of mutual strong interest in what the other is saying and doing. Mutual attentiveness often evoked through small talk and self-disclosure. Finally, Coordination refers to the sense of linguistic and non-linguistic synchrony arising from each other's echoing speech acts. Through these elements, effective rapport building is conducted.
In the context of research on language learning, previous studies have also demonstrated that a high level of rapport between learners and teachers can bring many positive impacts, such as increased learner achievement, motivation, and engagement\cite{xie2021conceptual}. Thus, from the persepective of the current study, it is vital to examine the extent to which dialogue agents are capable enough of building rapport with users.

However, no generalized method has been established for measuring engagement and rapport in evaluating dialogue systems. Therefore, this study attempts to evaluate them using questionnaire items based on these findings.
\section{Experimental Method}
In this study, we aim to develop and validate the scales of engagement and rapport as the measures of user experience in role-play and discussion tasks with multimodal dialogue systems for English language learning. To this end, we constructed a set of questionnaire items to tap into major sub-dimensions of the constructs based on relevant theories reviewed so far. We then examined the structural validity of the scales through Confirmatory Factor Analysis (CFA). In addition, we compared the scores of engagement and rapport between the dialogue with human interlocutors and the one with the system. 
 
\subsection{Dialogue Tasks}
For the conversation tasks, the role-play format from \cite{ikeda2017measuring} and the discussion task from \cite{ockey2021human} were adopted.
In the role-play, participants played a student role who requests an extension of a deadline from the teacher role (played by the interlocutor). The role-play card and the procedures were designed following the format of \cite{ikeda2017measuring}, which are presented in the appendix\footnote{\url{https://github.com/fumakurata/kurata_2024_IWSDS}}.
In the discussion task, both participants and interlocutors were first provided with two different views on the use of social networking serveice (SNS). They were then asked to summarise their assigned position. After the summarization phase, they were instructed to defend their own positions and convince the partner through a discussion. Tutors or AI agents played the examiner role, while learners took the student role.

\subsection{Participants}
\subsubsection{Learners}
A total of 74 Japanese learners of English were recruited in this study. 
They were undergraduate and graduate students from a private university in Japan, with their proficiency levels ranging from beginner to advanced English speakers. However, three participants did not complete the experiment, resulting in a total of 71 participants included in the analysis.

\subsubsection{Tutors}
A total of five English teachers served as tutors in this study. All were experienced instructors teaching English speaking skills at Japanese universities.
To standardize the tutors' behaviours in the dialogue tasks, researchers delivered a two-hour training session with the tutors. This included overall guidance and specific instructions for each role-play scenario and discussion task. During the session, they were required to converse with participants as naturally as possible following the role-play cards. To this end, they were not allowed to correct learners' mistakes or give explicit feedback on language issues during the dialogue tasks.

\subsection{Dialogue System Configuration}
\begin{figure}[t]
    \centering
    \includegraphics[width=\hsize]{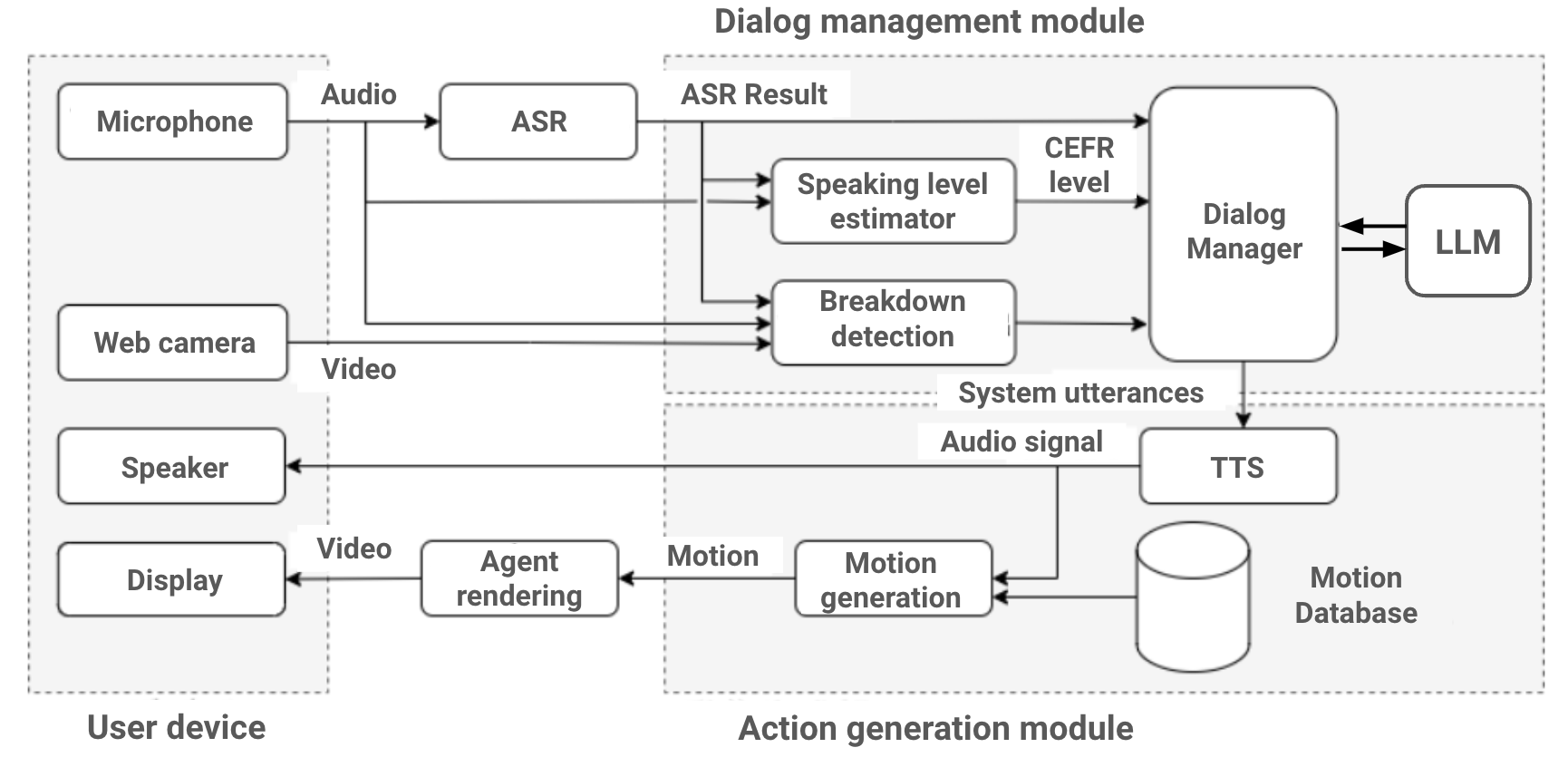}
    \caption{System architecture}
    \label{fig:system_architecture}
\end{figure}

In this study, we adopted a dialogue agent designed based on an interview agent for assessing English speaking ability \cite{saeki2021intella}. The architecture of this system is shown in Figure \ref{fig:system_architecture}.
All the components were kept as desigend by \cite{saeki2021intella}, except for the module of utterance generation. Specifically, this module in the original study adopted a scenario database. To handle more reciplocal interactions between speakers in the current dialogue task, compared to the interview task targeted in \cite{saeki2021intella}, this module was replaced with a Large Language Model (LLM) in the current study. 
For the LLM, OpenAI's GPT-4\footnote{\url{https://openai.com/research/gpt-4}} was used.

The agent transitioned phases in the order of Introduction, Main Task, and Closing. During the Introduction, greetings, self-introductions, and small talk took place. The Main Task involved  either the role-play or discussion, and the Closing wound down the dialogue.
Prompts given to the LLM varied depending on the phase and included situational settings, rules to adhere to within the dialogue, and the tasks the agent should perform in each phase such as greeting the user.
When all tasks are completed in each phase, a phase completion token is outputted so that the dialogue manager can transition to the next phase.
The dialogue ends when a phase completion token is outputted in the final phase.

\subsection{Experimental Design}
Learners were required to perform two tasks (role-play and discussion) in sequence with both tutors and AI agents. To mitigate order effects due to the difference in dialogue partners, the order of the condition of interlocutors (human tutors vs. virtual agent) was counterbalanced across participants. For instance, Group A completed both tasks with a human tutor in the first session and then completed both tasks with an AI agent in the second session.

After the completion of each task, participants responded to the questionnnaire items of engagement and rapport as shown in Table \ref{tab:engagement_question} and \ref{tab:rapport_question}. Their responses were collected online using Qualtrics\footnote{\url{https://www.qualtrics.com}}.

\subsection{Questionnaires}
We created the questionnaire items corresponding to the components of engagement and rapport based on \cite{oga2019acting, hiver2021engagement, zhao2014towards}. Since engagement and rapport are psychological constructs, the questionnaires were designed to include multiple items for each component so that the internal consistency of the items can be assessed, following the common practice in developing psychological scales. We modified existing questionnaires for dialogue tasks in this study. The questionnaire items for engagement are shown in Table \ref{tab:engagement_question}. Behavioral engagement includes the questions about learners' effort and ability to participate in conversations. Cognitive engagement contains the questions about concentration and perceived experience in dialogue tasks. Emotional engagement consists of the questions about learners' positive and negative feelings during dialogue tasks. For social engagement, the questions about attitudes towards the dialogue partner are adopted. Meanwhile, the items for rapport are shown in Table \ref{tab:rapport_question}. The items of Face management ask how much learners feel respected by their dialogue partners. Mutual attentiveness includes the questions regarding learners' perception of their dialogue partner's attitudes. Coordination contains the questions about the sense of synchrony in the conversation with the dialogue partner. All the items are designed with a 5-point likert scale ranging from 1 (\textit{strongly disagree}) to 5 (\textit{strongly agree}).

\begin{table*}[ht]
\centering
\caption{Engagement questionnaire items}
\label{tab:engagement_question}
\resizebox{\columnwidth}{!}{%
\begin{tabular}{l|l} 
\hline
\multicolumn{1}{c|}{}                                                                        & \multicolumn{1}{l}{Items}                \\ \hline
\multicolumn{1}{c|}{\multirow{3}{*}{\begin{tabular}[c]{@{}c@{}}Behavioral engagement\end{tabular}}} & Did you try to talk as much as possible? \\ 
\multicolumn{1}{c|}{}                                                                        & Were you able to maximize your English speaking ability?               \\
\multicolumn{1}{c|}{}                                                                        & Did you endeavor to communicate your ideas and feelings?                \\ \hline
\multirow{3}{*}{Cognitive engagement}                                                                 & Were you able to stay focused on the conversation?                 \\
                                                                                             & Were you able to immerse yourself in the conversation with the one you were talking to?                      \\
                                                                                             & Did you find yourself distracted during the conversation?            \\ \hline
\multirow{3}{*}{Emotional engagement}                                                                 & Did you enjoy the conversation with the one you were talking to?                       \\
                                                                                             & Would you like to do another conversation with the one you were talking to?              \\
                                                                                             & Did you feel stress or frustration in the conversation?               \\ \hline
\multirow{3}{*}{Social engagement}                                                                 & Did you have favorable feelings toward the one you were talking to?                  \\
                                                                                             & Did you feel a sense of familiarity with the one you were talking to? \\ 
                                                                                             & Did you feel that the one you were talking to understood your feelings?                 \\ \hline
\end{tabular}
}
\end{table*}

\begin{table*}[ht]
\centering
\caption{Rapport questionnaire items}
\label{tab:rapport_question}
\resizebox{\columnwidth}{!}{%
\begin{tabular}{l|l} 
\hline
                                      & Items                                      \\ \hline
\multirow{3}{*}{Face management}      & Did you feel that the one you were talking to respected you?                    \\
                                      & Did you find the one you were talking to friendly?                       \\
                                      & Did you feel that the one you were talking to cared about you?                    \\ \hline
\multirow{3}{*}{Mutual attentiveness} & Did you feel that the one you were talking to was listening to you?                \\
                                      & Did you feel that the one you were talking to was interested in you?                  \\
                                      & Did the one you were talking to seem to pay attention to your opinions and feelings?          \\ \hline
\multirow{3}{*}{Coordination}         & Did you feel that you had a good conversation with the one you were talking to?                     \\
                                      & Did you feel that you worked well with the one you were speaking with in facilitating the conversation?           \\
                                      & Did the one you were talking to provide feedback or ask questions to show understanding of your statements? \\ \hline
\end{tabular}
}
\end{table*}

\subsection{Analysis}
Three types of analyses were conducted in this study. The first analysis evaluated the reliability of the questionnaires by calculating Cronbach's alpha coefficient for each category. This coefficient ranges from 0 to 1, with values closer to 1 indicating higher internal consistency. Cronbach's alpha was calculated using the statistical analysis library in Python, pingouin\footnote{\url{https://pypi.org/project/pingouin/}}. 

\begin{figure}[t]
    \centering
    \includegraphics[width=0.9\hsize]{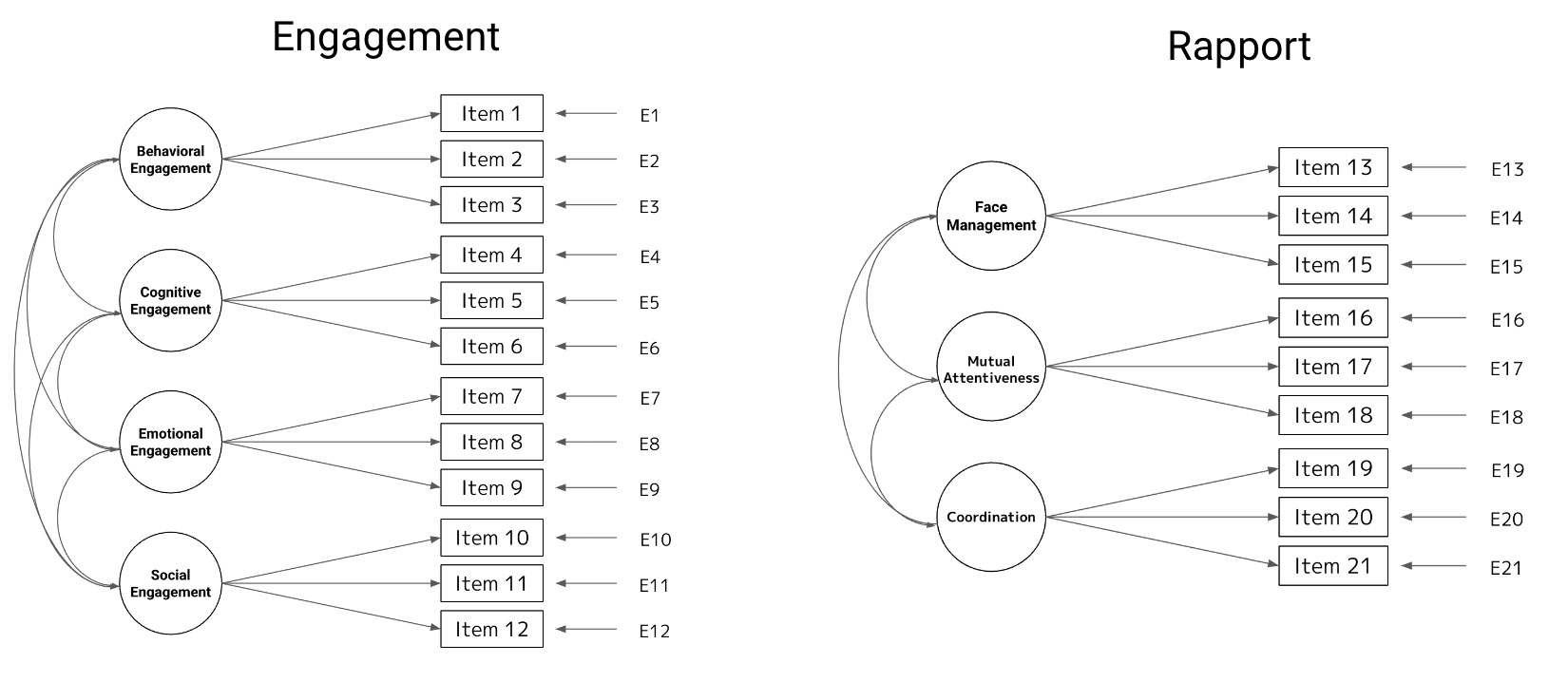}
    \caption{Hypothesis model}
    \label{fig:hypo_model}
\end{figure}

\begin{figure}[t]
    \centering
    \includegraphics[width=0.9\hsize]{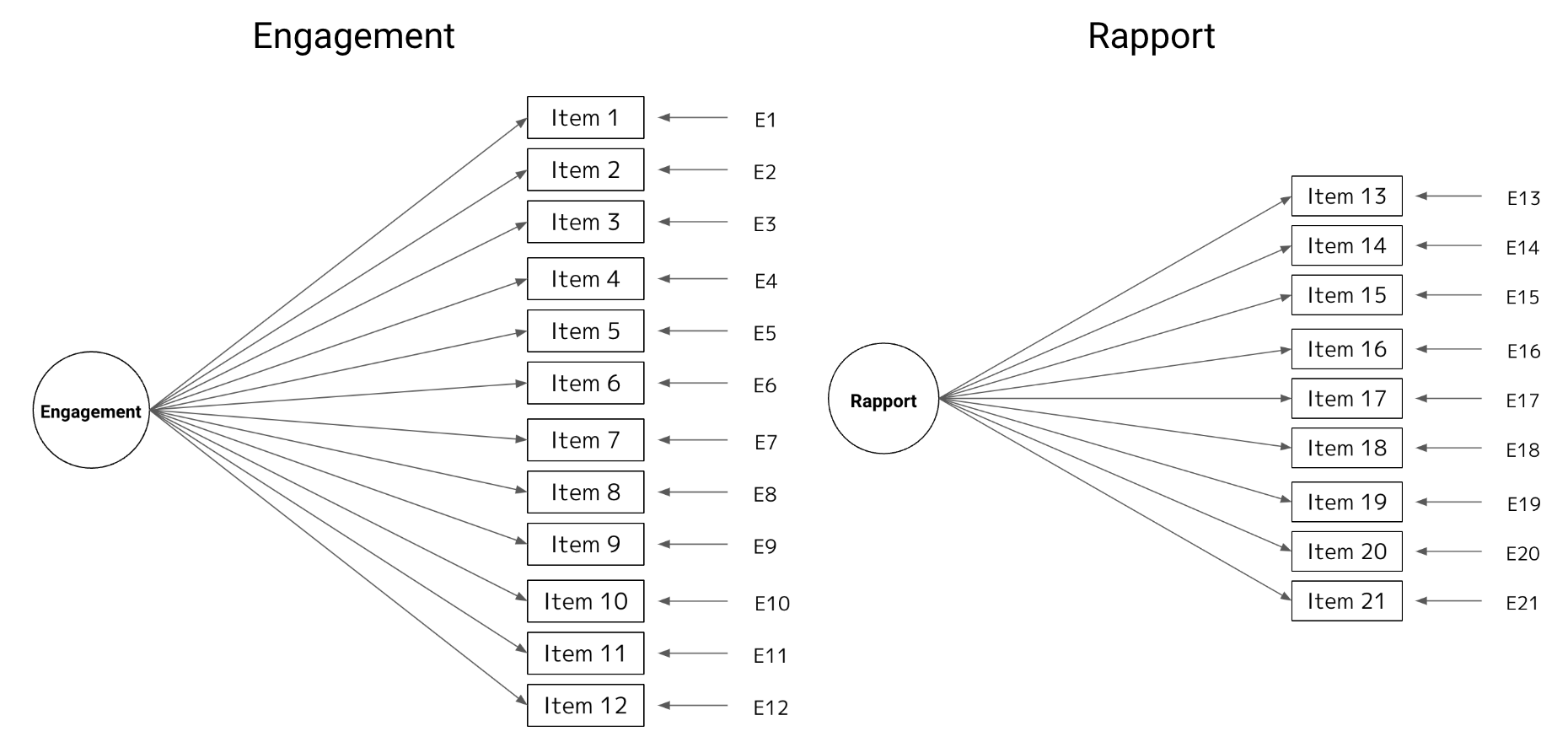}
    \caption{Baseline model}
    \label{fig:baseline_model}
\end{figure}

The second analysis involved CFA to validate the proposed factor structure of engagement and rapport. As shown in Figure \ref{fig:hypo_model}, we assumed four sub-constructs in the engagement model (i.e., Behavioral, Cognitive, Emotional, and Social) and three sub-constructs in the rapport model (i.e., Face Management, Mutual Attentiveness, and Coordination) based on prior research mentioned in Section \ref{related:engagement_rapport}. We estimated each of these components as latent variables from the three corresponding items shown in Tables \ref{tab:engagement_question} and \ref{tab:rapport_question}. For comparison, we also fitted a baseline model which assumes engagement and rapport as unified constructs, as shown in Figure \ref{fig:baseline_model}. 

To evaluate the model fit, we computed the following indices for the sake of the comparability with other studies: $\chi^2$, p-value, CFI (Comparative Fit Index), TLI (Tucker-Lewis Index), RMSEA (Root Mean Square Error of Approximation), SRMR (Standardized Root Mean Square Residual), AIC, and BIC were used. $\chi^2$ measures the model's absolute fit. 
 Ideally, if $\chi^2$ is small and the p-value is greater than 0.05, the model is considered to fit the data well; however, this metric is sensitive to sample size and requires careful interpretation.  CFI and TLI are indicators of relative fit, ranging from 0 to 1, with values above 0.90 indicating appropriate model fit. RMSEA measures the magnitude of the model's prediction error, and SRMR is an index of the average size of residuals. Smaller RMSEA and SRMR indicates a better fit, and less than 0.8 is desirable \cite{hu1999cutoff}. However, RMSEA can be unstable in small sample sizes. AIC and BIC consider both the model's fit and complexity, with smaller values indicating a better fit (although they cannot be used for comparing models with different observed variables). In the current study, due to the relatively small sample size (\textit{n} = 71), we partiucularly focus on the index of SRMR given its relative robustness to a small sample size.

After checking the fit indices, we investigated factor loadings, residuals, factor correlations for local fit. If necessary, adjustments were made to improve the local as well as global model fit. The analysis was conducted using the statistical analysis software, JASP\footnote{\url{https://jasp-stats.org/}}.

Finally, in the third analysis, a series of t-test was conducted to investigate the differences between human tutors and agents in the four factors of engagement and the three factors of rapport. The two reversed items ("Did you find yourself distracted during the interview?" and "Did you feel stress or frustration in the conversation?") had their scores reversed before CFA. 

\section{Results}
\subsection{Internal Consistency of Questionnaire Items}
\begin{figure}[ht]
  \centering
  \begin{subfigure}[b]{0.45\textwidth}
      \centering
      \includegraphics[width=\textwidth]{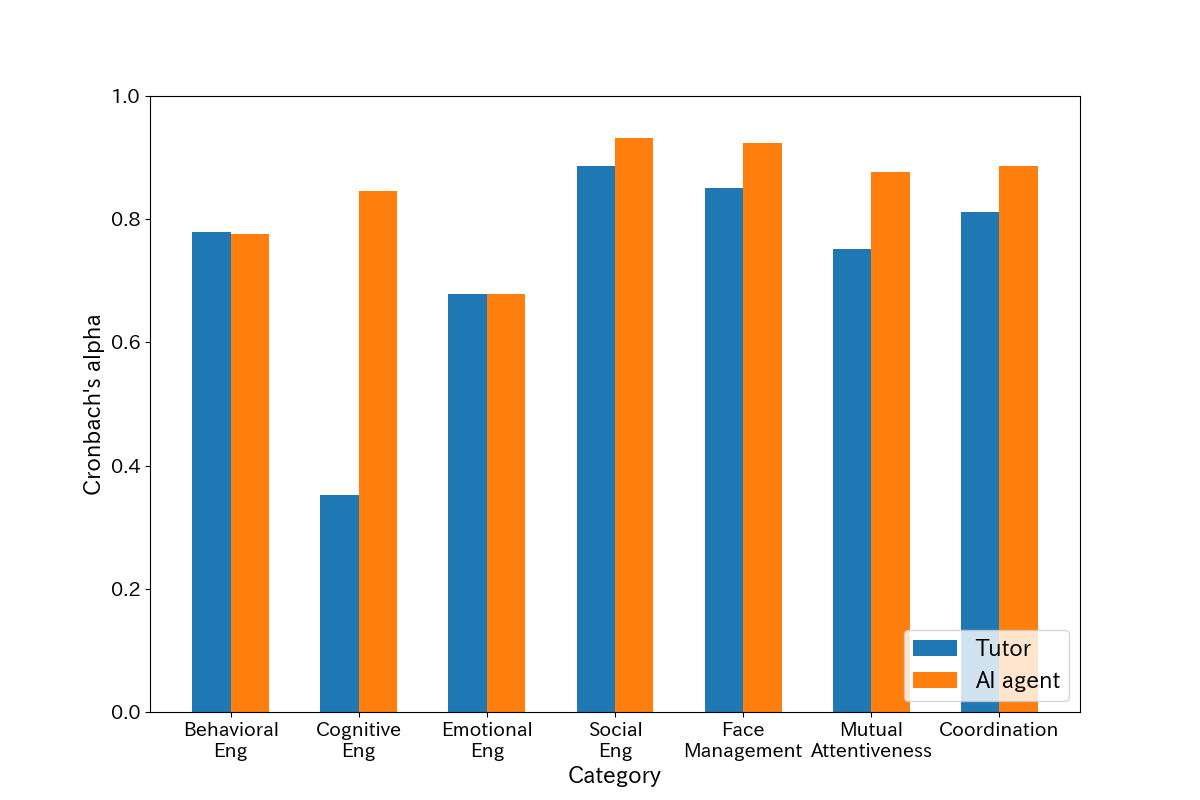}
      \caption{Crobach's alpha coefficient in role-play tasks}
      \label{fig:roleplay_chronbach}
  \end{subfigure}
  \hfill
  \begin{subfigure}[b]{0.45\textwidth}
      \centering
      \includegraphics[width=\textwidth]{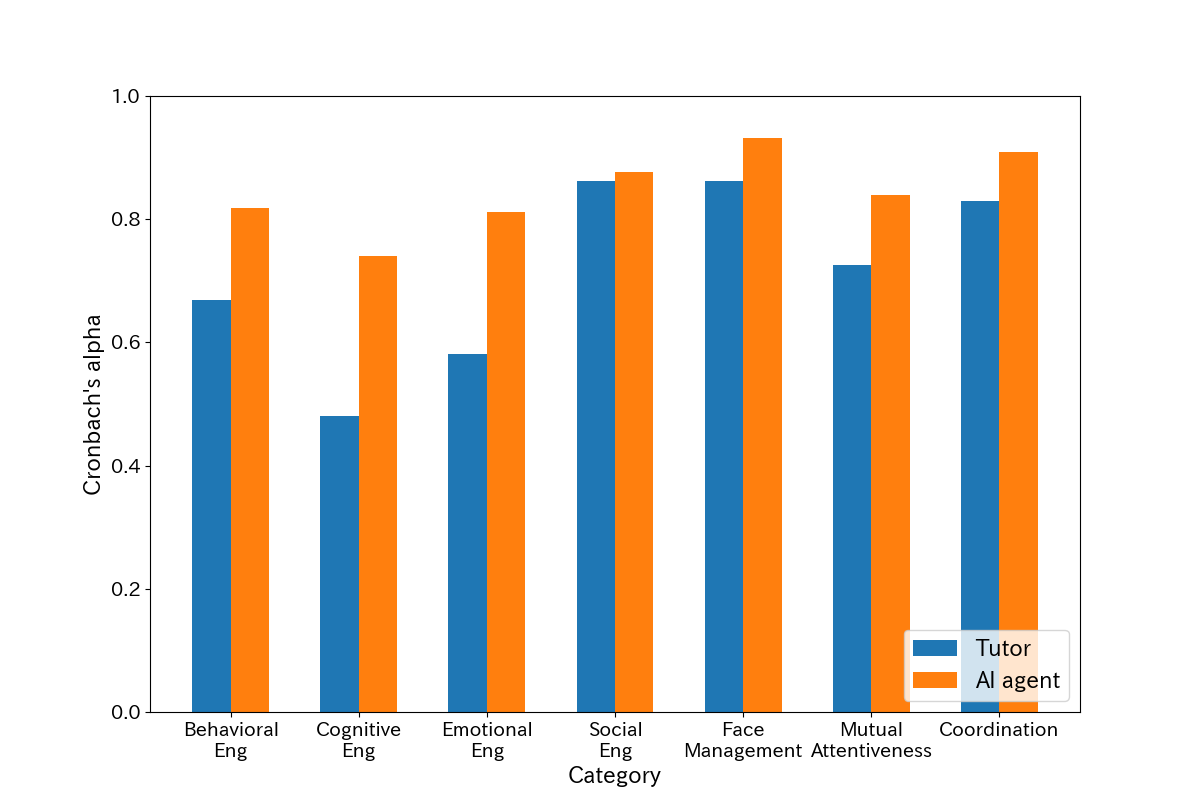}
      \caption{Crobach's alpha coefficient in discussion tasks}
      \label{fig:discussion_chronbach}
  \end{subfigure}
  \caption{Chrobach's alpha coefficient}
  \label{fig:chronbach}
\end{figure}

The Cronbach's alpha coefficients for the role-play task are shown in Figure \ref{fig:roleplay_chronbach}, and those for the discussion task are shown in Figure \ref{fig:discussion_chronbach}. In terms of engagement,  and social engagement showed high internal consistency in all cases($\alpha = 0.863-0.931$). However, the coefficients for behavioral and cognitive and emotional engagement were lower. Notably, the value for cognitive engagement was significantly lower when the dialogue partner was a human($\alpha = 0.238$). Meanwhile, for rapport, all items showed relatively high internal consistency ($\alpha = 0.726-0.932$).

\subsection{Confirmatory Factor Analysis}
We fitted both the baseline and hypothesized models for engagement and eapport for each interlocutor type (human tutor and AI). The fit indices for the final human-tutor models and the final agent models are respectively summarized in Table \ref{tab:CFA_result_human} and Table \ref{tab:CFA_result_intella}. In these final models, we allowed a pair of residual correlation between two reversed items ("Did you find yourself distracted during the conversation?" and "Did you feel stress or frustration in the conversation?"), which were both low in factor loadings. The fit indices for the measurement model of engagement (from the tutor data) indicated that the correlated four-factor model generally outperformed the baseline model in both role-play and discussion tasks as evidenced in increased CFI and TLI as well as decreased RMSEA and SRMR. Similarly, the hypothesized correlated three-factor model of rapport performed slighly better than the baseline uni-dimensional model (except in the discussion with tutors). These results showed that the hypothesized models generally capture the structure of the data better than the baseline models. This suggests that our engagement and rapport questionnaires can capture the hypothesized dimensionality of the psychological constructs, thus providing support for the structural validity of the scales.



\begin{table}[t]
\centering
\caption{CFA goodness-of-fit index (tutor)}
\label{tab:CFA_result_human}
\scalebox{0.7}{
\begin{tabular}{l|llll|llll}
\hline
                                               & \multicolumn{4}{c|}{Engagement}                                                    & \multicolumn{4}{c}{Rapport}                                                  \\ \cline{2-9} 
                                               & \multicolumn{2}{c|}{Roleplay}                    & \multicolumn{2}{c|}{Discussion} & \multicolumn{2}{c|}{Roleplay}              & \multicolumn{2}{c}{Discussion}  \\ \cline{2-9} 
                                               & baseline       & \multicolumn{1}{l|}{Hypothesis} & Baseline         & Hypothesis   & Baseline & \multicolumn{1}{l|}{Hypothesis} & Baseline       & Hypothesis     \\ \hline
$\chi^2$                          & 113.253        & \multicolumn{1}{l|}{67.760}     & 114.527          & 75.645       & 37.415   & \multicolumn{1}{l|}{31.718}     & 97.620         & 79.449         \\
df                                             & 53             & \multicolumn{1}{l|}{47}         & 53               & 47           & 27       & \multicolumn{1}{l|}{24}         & 27             & 24             \\
p                                              & \textless .001 & \multicolumn{1}{l|}{0.025}      & \textless .001   & 0.005        & 0.088    & \multicolumn{1}{l|}{0.134}      & \textless .001 & \textless .001 \\
Comparative Fit Index(CFI)                     & 0.887          & \multicolumn{1}{l|}{0.961}      & 0.869            & 0.939        & 0.977    & \multicolumn{1}{l|}{0.983}      & 0.846          & 0.879          \\
Tucker-Lewis Index(TLI)                        & 0.859          & \multicolumn{1}{l|}{0.943}      & 0.837            & 0.915        & 0.970    & \multicolumn{1}{l|}{0.975}      & 0.795          & 0.819          \\
AIC                                            & 563.906        & \multicolumn{1}{l|}{530.413}    & 651.560          & 624.678      & -16.760  & \multicolumn{1}{l|}{-16.456}    & 85.125         & 72.954         \\
BIC                                            & 647.625        & \multicolumn{1}{l|}{627.708}    & 735.279          & 721.973      & 44.333   & \multicolumn{1}{l|}{51.424}     & 146.218        & 140.835        \\
Root mean square error of approximation(RMSEA) & 0.127          & \multicolumn{1}{l|}{0.079}      & 0.128            & 0.093        & 0.074    & \multicolumn{1}{l|}{0.067}      & 0.192          & 0.180          \\
RMSEA 90\% CI lower bound                      & 0.094          & \multicolumn{1}{l|}{0.029}      & 0.096            & 0.051        & 0.000    & \multicolumn{1}{l|}{0.000}      & 0.152          & 0.137          \\
RMSEA 90\% CI upper bound                      & 0.159          & \multicolumn{1}{l|}{0.118}      & 0.160            & 0.130        & 0.126    & \multicolumn{1}{l|}{0.125}      & 0.234          & 0.225          \\
Standardized root mean square residual(SRMR)   & 0.058          & \multicolumn{1}{l|}{0.051}      & 0.065            & 0.053        & 0.037    & \multicolumn{1}{l|}{0.033}      & 0.067          & 0.076          \\ \hline
\end{tabular}
}
\end{table}

\begin{table}[t]
\centering
\caption{CFA goodness-of-fit index (AI agent)}
\label{tab:CFA_result_intella}
\scalebox{0.7}{
\begin{tabular}{l|llll|llll}
\hline
                                               & \multicolumn{4}{c|}{Engagement}                                                    & \multicolumn{4}{c}{Rapport}                                                  \\ \cline{2-9} 
                                               & \multicolumn{2}{c|}{Roleplay}                    & \multicolumn{2}{c|}{Discussion} & \multicolumn{2}{c|}{Roleplay}              & \multicolumn{2}{c}{Discussion}  \\ \cline{2-9} 
                                               & baseline       & \multicolumn{1}{l|}{Hypothesis} & Baseline         & Hypothesis   & Baseline & \multicolumn{1}{l|}{Hypothesis} & Baseline       & Hypothesis     \\ \hline
$\chi^2$                         & 140.956        & \multicolumn{1}{l|}{82.368}     & 145.998          & 72.527       & 40.141   & \multicolumn{1}{l|}{34.992}     & 95.219         & 79.548         \\
df                                             & 53             & \multicolumn{1}{l|}{47}         & 53               & 47           & 27       & \multicolumn{1}{l|}{24}         & 27             & 24             \\
p                                              & \textless .001 & \multicolumn{1}{l|}{0.001}      & \textless .001   & 0.010        & 0.050    & \multicolumn{1}{l|}{0.069}      & \textless .001 & \textless .001 \\
Comparative Fit Index(CFI)                     & 0.871          & \multicolumn{1}{l|}{0.948}      & 0.853            & 0.960        & 0.981    & \multicolumn{1}{l|}{0.984}      & 0.897          & 0.916          \\
Tucker-Lewis Index(TLI)                        & 0.840          & \multicolumn{1}{l|}{0.927}      & 0.816            & 0.943        & 0.974    & \multicolumn{1}{l|}{0.976}      & 0.862          & 0.874          \\
AIC                                            & 1871.365       & \multicolumn{1}{l|}{1824.777}   & 1072.617         & 1011.146     & 1175.659 & \multicolumn{1}{l|}{1176.511}   & 725.727        & 716.056        \\
BIC                                            & 1955.084       & \multicolumn{1}{l|}{1922.073}   & 1156.336         & 1108.441     & 1236.751 & \multicolumn{1}{l|}{1244.391}   & 786.820        & 783.937        \\
Root mean square error of approximation(RMSEA) & 0.153          & \multicolumn{1}{l|}{0.103}      & 0.157            & 0.087        & 0.083    & \multicolumn{1}{l|}{0.080}      & 0.189          & 0.181          \\
RMSEA 90\% CI lower bound                      & 0.123          & \multicolumn{1}{l|}{0.065}      & 0.127            & 0.044        & 0.003    & \multicolumn{1}{l|}{0.000}      & 0.148          & 0.138          \\
RMSEA 90\% CI upper bound                      & 0.184          & \multicolumn{1}{l|}{0.139}      & 0.188            & 0.126        & 0.134    & \multicolumn{1}{l|}{0.135}      & 0.230          & 0.225          \\
Standardized root mean square residual(SRMR)   & 0.064          & \multicolumn{1}{l|}{0.056}      & 0.075            & 0.050        & 0.028    & \multicolumn{1}{l|}{0.027}      & 0.046          & 0.040          \\ \hline
\end{tabular}
}
\end{table}

\begin{figure}[th]
    \centering
    \includegraphics[width=0.8\hsize]{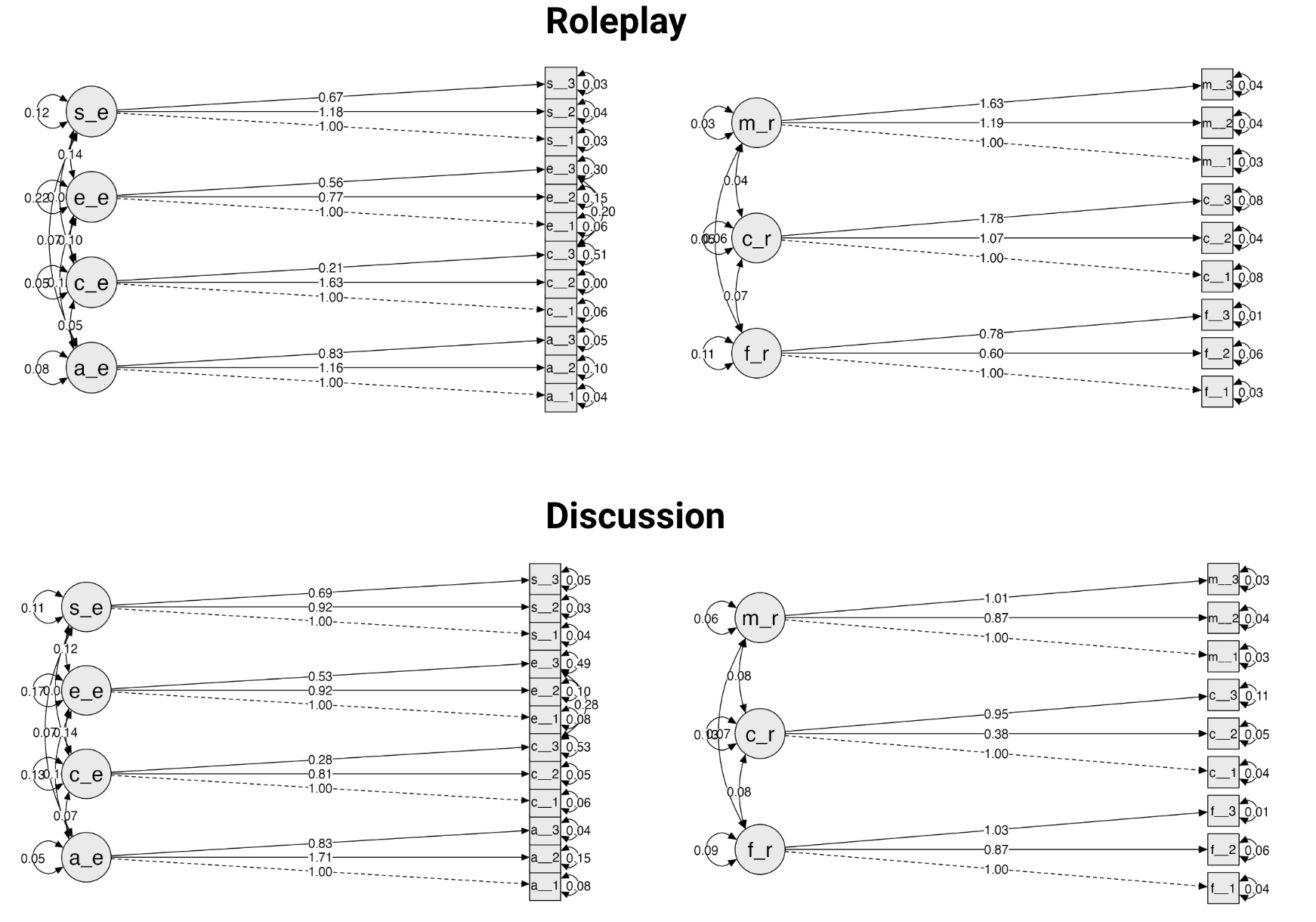}
    \caption{Path diagram and factor loadings (tutor)}
    \label{fig:path_diagram_human}
\end{figure}

To further examine the validity of the items, the factor loadings for the hypothesized models are discussed. The path diagrams of the models with human data and the agent data are res[ectively visualized in Figure \ref{fig:path_diagram_human} and Figure \ref{fig:path_diagram_intella}. Detailed statistics on loadings are reported in the appendix\footnote{\url{https://github.com/fumakurata/kurata_2024_IWSDS}}. The factor loadings between each latent variable and the corresponding observed variables are generally high, confirming the validity of the questionnaire items. However, the two reversed items of engagement ("Did you find yourself distracted during the conversation?" and "Did you feel stress or frustration in the conversation?") showed lower factor loadings compared to other items. One possible reason for this could be that during the online questionnaire collection, learners might not have realized that these were reverse items and then inadvertently provided incorrect responses, subsequently adding noise to the data. Therefore, constructing a model with higher fit might have been achieved without the reversed items in the questionnaire.

\begin{figure}[ht]
    \centering
    \includegraphics[width=0.8\hsize]{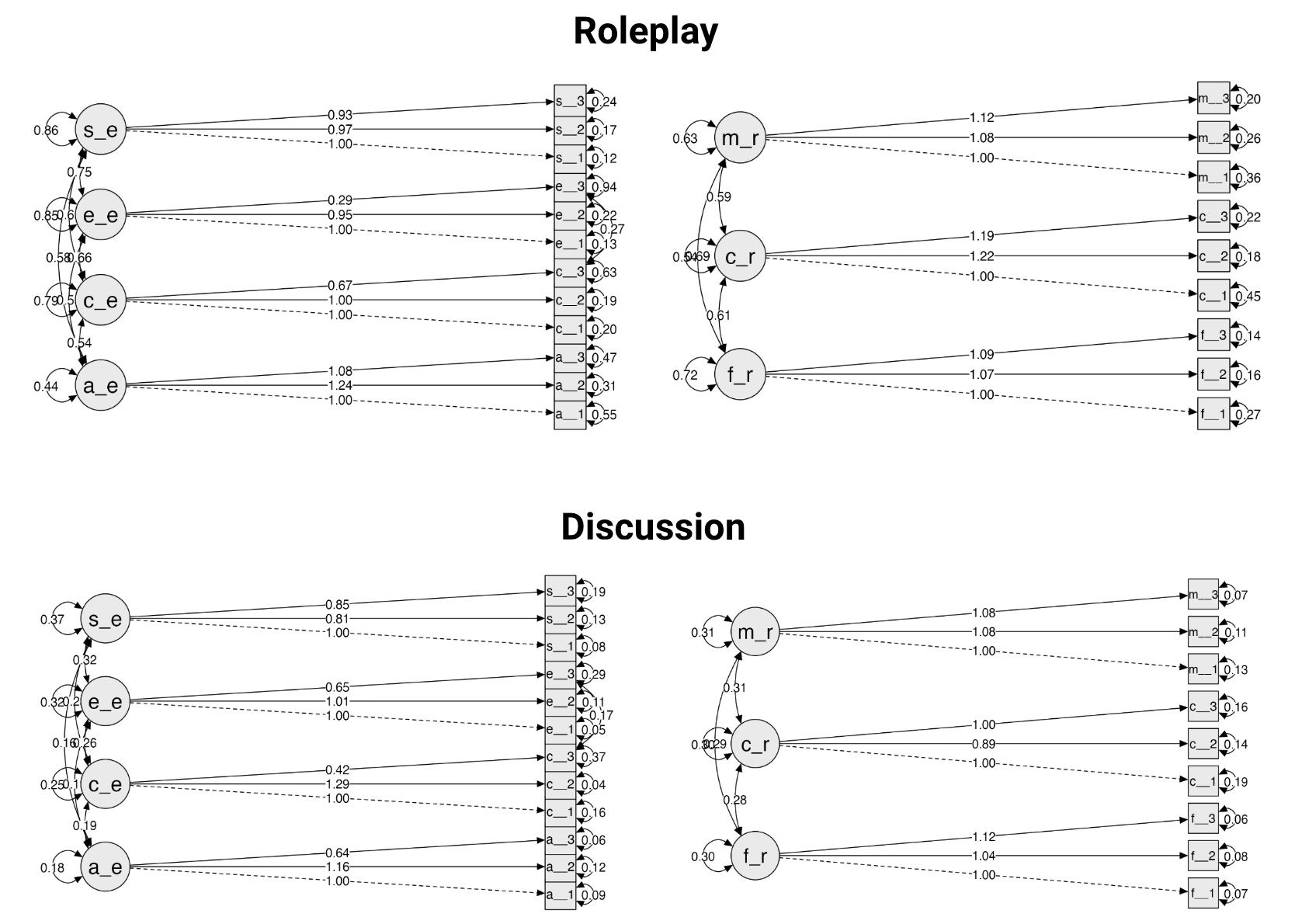}
    \caption{Path diagram and factor loadings (AI agent)}
    \label{fig:path_diagram_intella}
\end{figure}

\subsection{Average Scores and t-tests}
\begin{figure}[ht]
  \centering
  \begin{subfigure}[b]{0.49\textwidth}
      \centering
      \includegraphics[width=\textwidth]{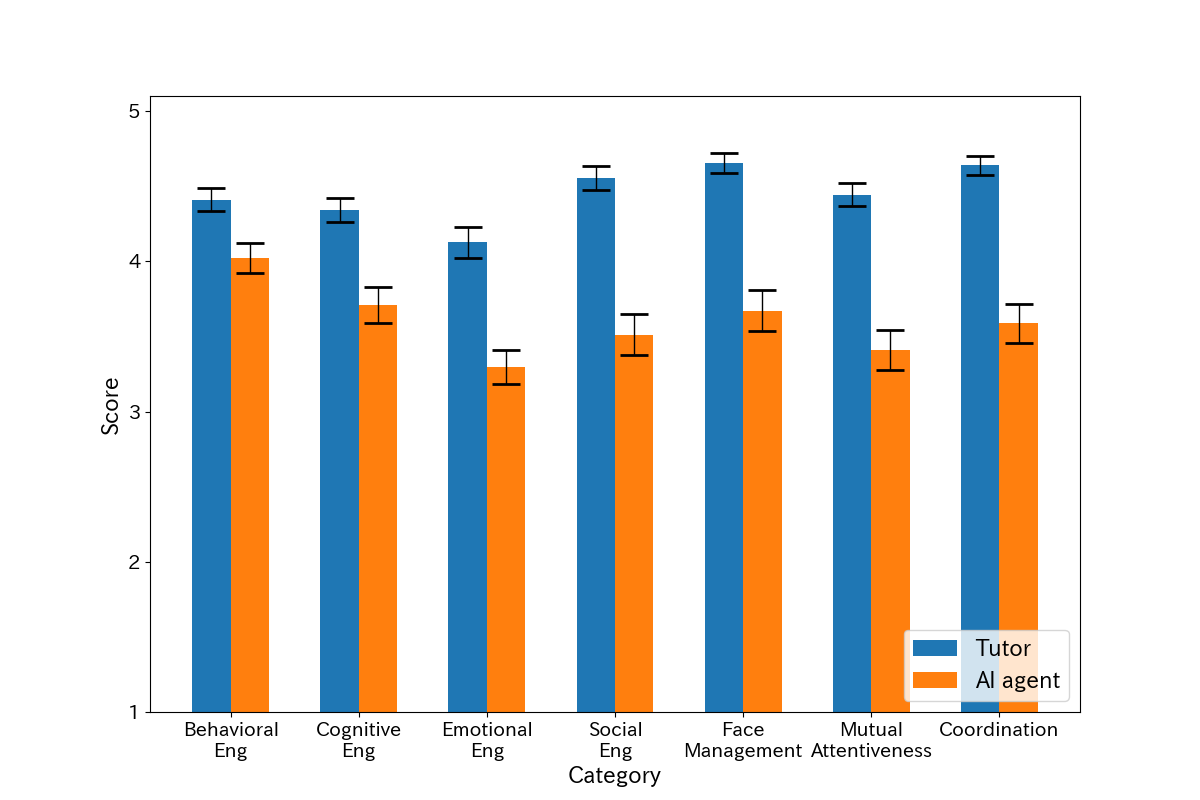}
      \caption{Average score in role-play tasks}
      \label{fig:roleplay_mean}
  \end{subfigure}
  \hfill
  \begin{subfigure}[b]{0.49\textwidth}
      \centering
      \includegraphics[width=\textwidth]{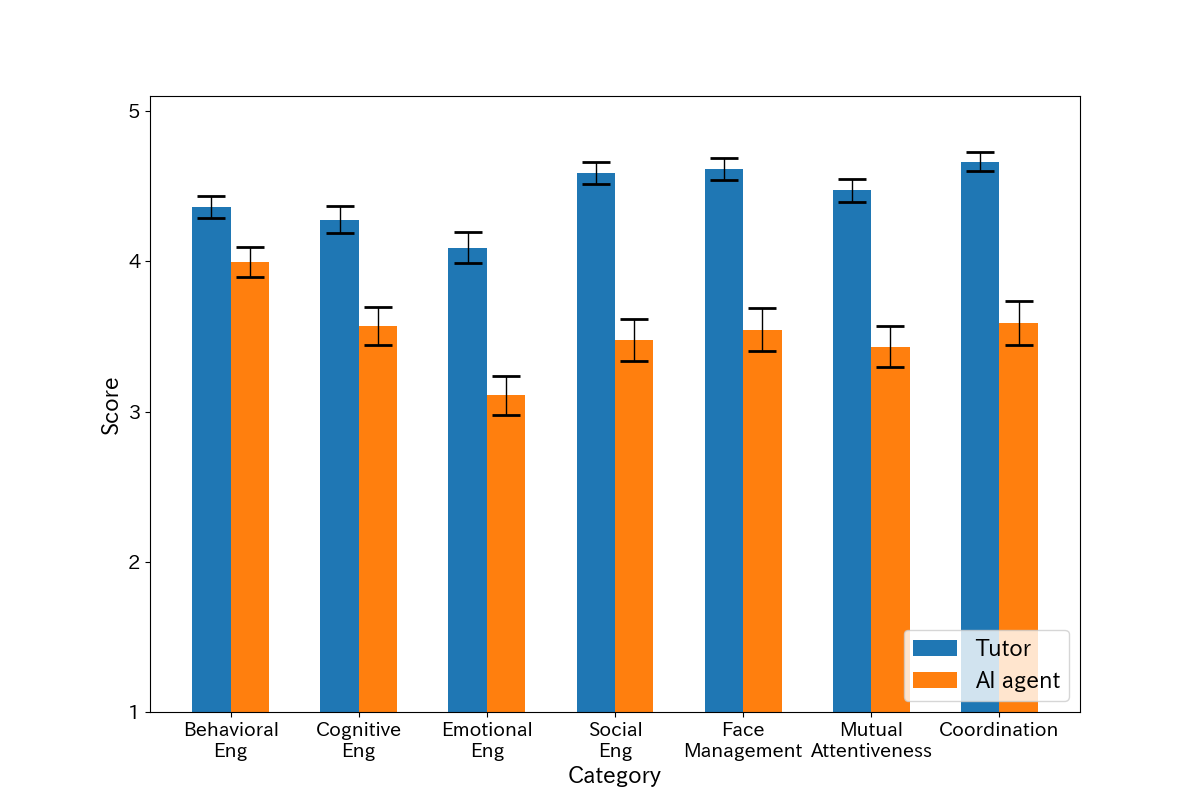}
      \caption{Average score in discussion tasks}
      \label{fig:discussion_mean}
  \end{subfigure}
  \caption{Average scores}
  \label{fig:chronbach}
\end{figure}

The average scores for the role-play task and the discussion task are respectively presented in Figure \ref{fig:roleplay_mean} and Figure \ref{fig:discussion_mean}. In both tasks, human tutors commonly had significantly higher average scores across all items ($p<0.05$). When averaging the scores of all items, for the role-play task, the score was 4.45 for human tutors and 3.60 for the agent, and for the discussion task, it was 4.44 for human tutors and 3.53 for the agent. Thus, it is demonstrated that human tutors scored higher in engagement and rapport compared to our AI agent.



\section{Discussion}
In this study, we developed and validated the questionnaires to measure engagement and rapport as indicators of user's psychological state and social relationship with the system to evaluate the user’s QoE in English conversation tasks. The results of confirmatory factor analyses indicated that our engagement and rapport questionnaires can capture the respective constructs. 
A follow-up analysis revealed the differences in conversation quality between humans and between human and agent in terms of engagement and rapport. However, more research is necessary to examine whether these questionnaires can be applied for dialogue tasks other than English learning. In dialogue tasks such as counseling, enhancing patients' willingness to talk and building trust are crucial for providing meaningful support to patients. In such situations, user's engagement and rapport may serve as effective evaluation metrics to optimize the behaviors of counseling agents. As illustrated by this example, engagement and rapport as psychological and social constructs would function as useful evaluation metrics for various tasks beyond task-oriented dialogue. 
Additionally, the engagement and rapport questionnaire data can be used for automatic estimation of these factors in future research. If user engagement and rapport can be automatically estimated with assistance of machine learning techniques, the holistic evaluation of dialogues could be accomplished without extensive data collection of user responses, substantially reducing evaluation costs. Furtheremore, the knowledge about dialogue features that can contribute to user's engagement and rapport might also help dialogue system developers identify which components of the system could enhance user experience quality.
\section{Conclusion}
This study proposed an evaluation method using engagement and rapport as metrics for assessing the quality of dialogues in English conversation tasks. Specifically, questionnaire items based on theoretical models of engagement and rapport were developed, drawing from insights in the field of educational psychology. In the experiment, dialogues conducted by human tutors and an AI agent were implemented, and learners' responses to questions regarding engagement and rapport were collected. The analysis using Cronbach's alpha coefficient and confirmatory factor analysis supported the validity of questionnaire items and their abilities to capture the assumed theoretical models of engagement and rapport. It was also shown that human tutors scored higher in engagement and rapport compared to our AI agent.
Future work will focus on developing methods for automatically estimating engagement and rapport scores and systems that provide evidence of system behaviors that influence these scores. The goal is to establish a workflow for improving the quality of dialogue experiences.

\section{Acknowledgement}
The research presented in this study was achieved through funding and support from the project "Technological Development for Next-Generation Artificial Intelligence Evolving with Humans (JPNP20006) / Development of an Online Language Learning Support AI System that Grows with Humans" by  New Energy and Industrial Technology Development Organization (NEDO), and  "Beyond 5G Seed Creation Program / Research and Development of an XR Communication Infrastructure for Realizing High-Immersion Interaction Experiences with Conversational AI Agents (JPJ012368C06301)" by the National Institute of Information and Communications Technology (NICT).

\bibliographystyle{unsrt}
\bibliography{refs}
\end{document}